\documentclass[letterpaper, 10 pt, conference]{ieeeconf}  

\IEEEoverridecommandlockouts                              

\usepackage{booktabs}
\usepackage{multirow}
\usepackage{graphicx}
\usepackage{color}
\usepackage{hyperref}

\usepackage{tikz}
\usetikzlibrary{positioning, arrows.meta, backgrounds, fit}
\usepackage{amssymb}
\usepackage{threeparttable}
\usepackage{booktabs}
\usepackage{tabularx}
\usepackage{makecell}

\title{\LARGE \bf
PACE: Persona Adaptation through Conversational Elicitation \\in Human-Robot Interaction
}

\author{Peizhen Li$^{1, *}$, Longbing Cao$^{1}$, Megani Rajendran$^{2}$, Timothy Liu$^{2}$, Aik Beng Ng$^{2}$, and Simon See$^{2}$ \\[2pt]
\includegraphics[height=0.8cm]{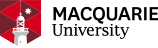}%
\hspace{1.5cm}%
\includegraphics[height=0.75cm]{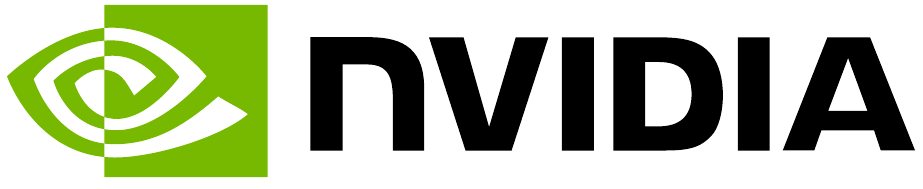}%
\vspace{-8pt}%
\thanks{$^{1}$Macquarie University, Australia. $^{2}$NVIDIA, Singapore.}%
\thanks{$^{*}$ Correspondence: \texttt{peizhen.li1@hdr.mq.edu.au}.}%
}

\begin{document}

\maketitle
\thispagestyle{empty}
\pagestyle{empty}

\begin{abstract}
Equipping humanoid robots with coherent and adaptable personas is crucial for fostering natural, engaging, and trustworthy human-robot interaction (HRI). However, existing approaches often rely on static, hard-coded identities that lack the flexibility to adapt to individual user contexts. In this paper, we present PACE (Persona Adaptation through Conversational Elicitation), a novel framework for the interactive generation and deployment of structured personas on the Ameca humanoid robot. Our system introduces an Interactive Persona Elicitation Pipeline, enabling the robot to dynamically synthesize a tailored, psychologically grounded identity through user Q\&A. This elicitation process feeds into a persona prompt compilation phase, generating a structured persona prompt built upon multi-perspective dimensions. We detail the Embodied System Integration required to translate this structured specification into expressive, multimodal humanoid behaviors. Through a comprehensive empirical HRI evaluation, we assess the impact of dynamically generated personas on user trust, perceived anthropomorphism, persona consistency, personal relevance, and interaction quality compared to a generic baseline. These contributions establish a scalable pathway for deploying personalized, interactive, and reliable identities in embodied humanoid assistants.
\textit{Video demo is available at: \url{https://lipzh5.github.io/PACE/}}

\end{abstract}

\section{INTRODUCTION}
The deployment of humanoid robots in social, collaborative, and service-oriented environments relies not only on their physical dexterity but also on their social intelligence~\cite{cao2025humanoid,li2025ugotme}. In human-robot interaction (HRI), the perception of a robot as a trustworthy and engaging partner is heavily influenced by its assigned persona~\cite{pruitt2003personas}---the coherent set of traits, values, and communicative styles it exhibits during an interaction~\cite{chien2022influence,zhu2025trust,li2026realm}. Recent advancements have successfully enabled real-time, realistic facial expression shadowing and physical mimicry in humanoid robots~\cite{li2026vividface,li2025x2c}. However, transitioning these robots from purely reactive imitators to autonomous, socially aware agents requires a fundamental shift toward dynamic cognitive persona adaptation. While recent breakthroughs in Large Language Models (LLMs) have enabled robots to engage in highly sophisticated, open-ended dialogue, these systems typically rely on static, developer-written ``system prompts'' or hard-coded character profiles that remain fixed throughout deployment.

\begin{figure}[t]
    \centering
    \includegraphics[width=\linewidth]{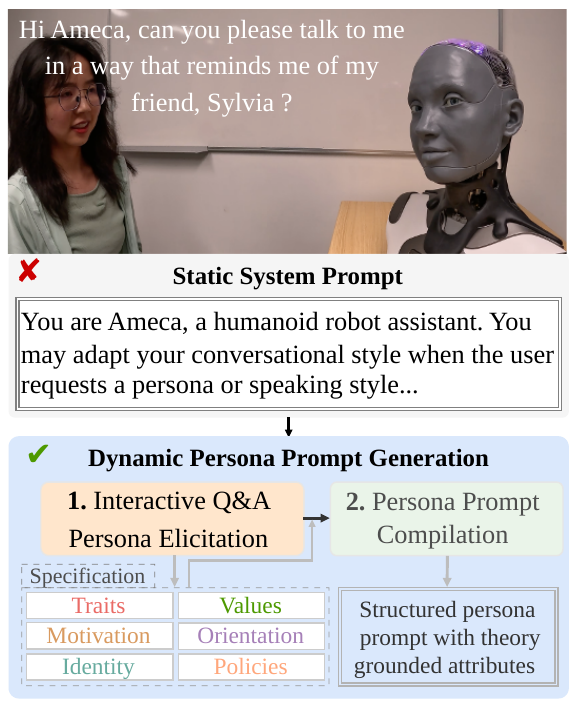}
    \caption{\textbf{Motivation and overview of the proposed PACE framework.} When a user asks the Ameca robot to adopt a specific, familiar conversational style, a traditional Static System Prompt offers limited adaptability. In contrast, our proposed Dynamic Persona Prompt Generation pipeline leverages Interactive Q\&A to build a comprehensive Persona Elicitation Specification. By extracting key psychological dimensions---Traits, Motivation, Values, Orientation, and Identity---the system performs a Persona Prompt Compilation to generate a structured persona prompt with theory-grounded attributes, which is subsequently mapped directly to physical hardware behaviors.}
    \label{fig:motivation}
\end{figure}

This reliance on static identity representations introduces a critical limitation in physical HRI: inflexibility. A rigid, pre-defined persona cannot adapt to the diverse preferences, cultural boundaries, and situational contexts of individual users. In a physically embodied agent, this lack of adaptation is particularly jarring; a mismatch between a user's expectations and a robot's embedded persona can induce cognitive dissonance, severely degrade user trust, break interactional immersion, and ultimately reduce the overall effectiveness of the embodied agent. For example, a user interacting with a robotic physical therapy assistant may require a firm, highly conscientious motivational coach, whereas another user dealing with high cognitive load might prefer a gentle, empathetic, and highly agreeable companion.

To address this critical gap, we introduce \textbf{PACE} (Persona Adaptation through Conversational Elicitation), a novel framework that shifts humanoid identity generation from static prompt engineering to dynamic, interactive elicitation. Fig.~\ref{fig:motivation} illustrates the motivation and high-level overview of this approach, contrasting traditional static prompts with our dynamic pipeline. Implemented on the highly expressive Ameca humanoid robot, our system allows the robot to actively interview the user through a natural language Q\&A phase before executing its primary task. By parsing the user's unstructured verbal responses, the framework dynamically compiles a structured, psychologically grounded persona specification---encompassing specific behavioral traits, values, and situational policies. 

Crucially, this compiled identity is not merely text-based; it is systematically integrated into Ameca’s physical embodiment. In this context, multimodal humanoid behavior refers primarily to the real-time synchronization of verbal communication and facial expressions. Our system dynamically infers appropriate facial affect based on the conversational context and seamlessly blends these macro-expressions with low-level speech visemes to match the synthesized persona, creating a highly congruent and believable physical presence.

This work bridges the gap between structured psychological AI frameworks and physical humanoid embodiment, demonstrating that dynamic persona generation significantly enhances the quality of real-world human-robot interaction. The primary contributions of this paper are as follows:
\begin{itemize}
    \item \textbf{Interactive Persona Elicitation Pipeline:} We propose a novel, conversational Q\&A interaction framework enabling a humanoid robot to dynamically synthesize a tailored, user-aligned identity prior to task execution, bypassing the need for exhaustive, fatigue-inducing psychological surveys.
    \item \textbf{Embodied System Integration:} We detail an integration architecture that translates compiled, structured psychological parameters into the multimodal physical constraints of the Ameca robot. This includes inferring contextually appropriate facial affect and dynamically blending these expressions with speech visemes during verbal delivery.
    \item \textbf{Empirical HRI Evaluation:} We present an in-person user study showing that dynamically compiled personas significantly improve multiple embodied HRI metrics, including perceived trust, anthropomorphism, persona consistency, personal relevance, and overall interaction quality, compared to a non-tailored static baseline.
\end{itemize}

\section{RELATED WORK}

\textbf{From Static to Adaptive Personas in HRI:} 
The design and integration of personas in social robots have been extensively explored to improve user engagement and establish social presence. Foundational works have largely focused on defining distinct personality traits within predefined scenarios~\cite{mohammadi2019designing} or tailoring expressive frameworks to specific, hard-coded character archetypes~\cite{whittaker2021designing}. While these methodologies successfully establish an initial social identity, they predominantly rely on static formulations and lack the capability to dynamically adapt in real-time to unfolding natural language conversations. To address the limitations of rigid behaviors, research has transitioned toward adaptive personas by integrating computational models to modify companion robot behavior over time~\cite{duque2013different}, alongside broader user-centered adaptation strategies, such as matching robot personality to post-stroke rehabilitation patients~\cite{tapus2008user}. Despite these advancements toward behavioral flexibility, a significant architectural gap remains: there is currently a lack of structured, multi-perspective pipelines dedicated to both the real-time conversational formulation of these dynamic personas and their rigorous, standardized evaluation during physically embodied human-robot interaction.

\textbf{LLM-Driven Adaptation and Embodiment:} 
Recent breakthroughs in LLMs have introduced powerful new methodologies for simulating human behavior and generating dynamic conversational agents~\cite{park2023generative, barmann2024incremental, xi2025rise, wang2024survey,chen2024persona,rajendran2025advancing}. However, these models typically rely on static ``system prompts'' that fail to interactively elicit user preferences prior to task execution. Furthermore, their textual outputs are rarely mapped systematically to the physical kinematic constraints of highly expressive humanoid robots. This physical mapping is a critical requirement for authentic interaction, a concept increasingly emphasized in recent humanoid robotics literature focusing on non-verbal and emotional feedback mechanisms~\cite{saood2024designing}. 

Our proposed PACE framework directly addresses these critical limitations. By shifting from static prompt engineering to an Interactive Persona Elicitation Pipeline, PACE dynamically compiles a psychologically grounded identity tailored to the user's immediate cognitive and emotional context. Crucially, our system bridges the persistent gap between text-based generation and physical HRI by translating these elicited traits into the multimodal embodiment of the Ameca robot, thereby demonstrably enhancing user trust, behavioral coherence, and overall interaction quality.

\section{SYSTEM ARCHITECTURE AND METHODOLOGY}

\begin{figure}[h]
    \centering
    \includegraphics[width=1\linewidth]{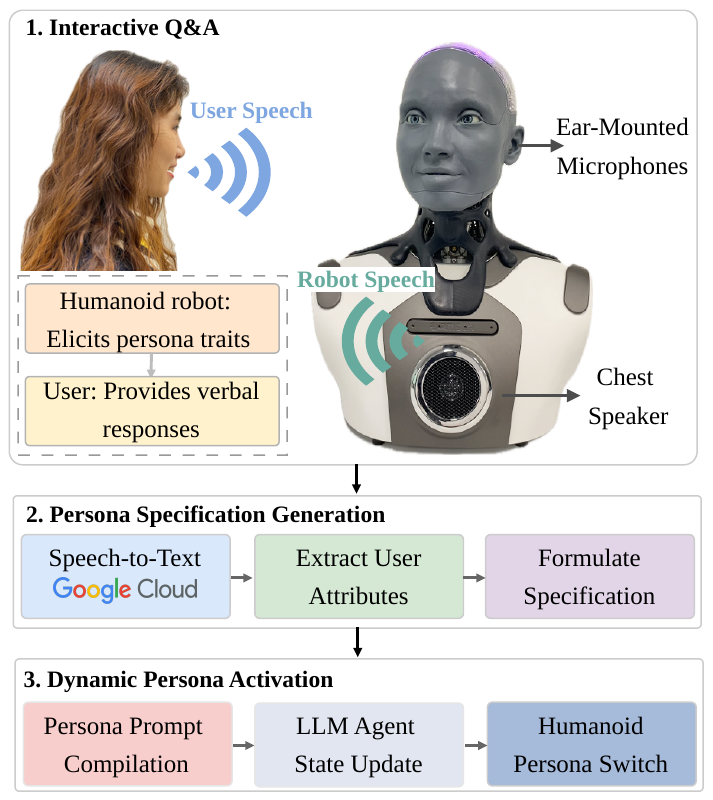}
    \caption{\textbf{Overview of the end-to-end system architecture for dynamic persona generation and deployment.} The pipeline transitions from (1) Interactive Q\&A for initial trait elicitation, to (2) Persona Specification Generation for attribute extraction and transcription, and concludes with (3) Dynamic Persona Activation, which compiles the prompt and executes the physical persona switch on the humanoid hardware.}
    \label{fig:system_overview}
\end{figure}

To overcome the inherent limitations of static character profiles, we introduce a novel pipeline for dynamic, data-driven persona generation. As illustrated in Fig.~\ref{fig:system_overview}, the proposed end-to-end PACE architecture operates across three continuous stages designed to seamlessly translate unstructured human conversational input into expressive physical behaviors. 

First, the \textbf{Interactive Q\&A} module handles the initial user engagement. The humanoid robot elicits persona traits via its chest speaker, and the user's unstructured verbal responses are captured by ear-mounted microphones. Second, the \textbf{Persona Specification Generation} stage processes this audio. It leverages Google Cloud Speech-to-Text to transcribe the interaction, extracts underlying psychological attributes via multi-perspective analysis, and formulates a structured specification profile. Finally, the \textbf{Dynamic Persona Activation} module closes the loop. It compiles the generated specification into a system prompt and updates the LLM agent's state. This state update is then physically executed on the hardware, interfacing directly with the robot's vocal generation and facial actuation endpoints to perform the humanoid persona switch.

\subsection{Interactive Q\&A}

\textbf{Adaptive Question Set Design:} 
While exhaustive demographic and life-story interviews can successfully capture intricate personal idiosyncrasies, reproducing a full two-hour psychometric interview in a physical human-robot deployment introduces severe user fatigue and degrades the interaction experience. To resolve this tension, our elicitation pipeline employs a dialogue management strategy that balances a structured psychological foundation with dynamic, multi-tier branching. 

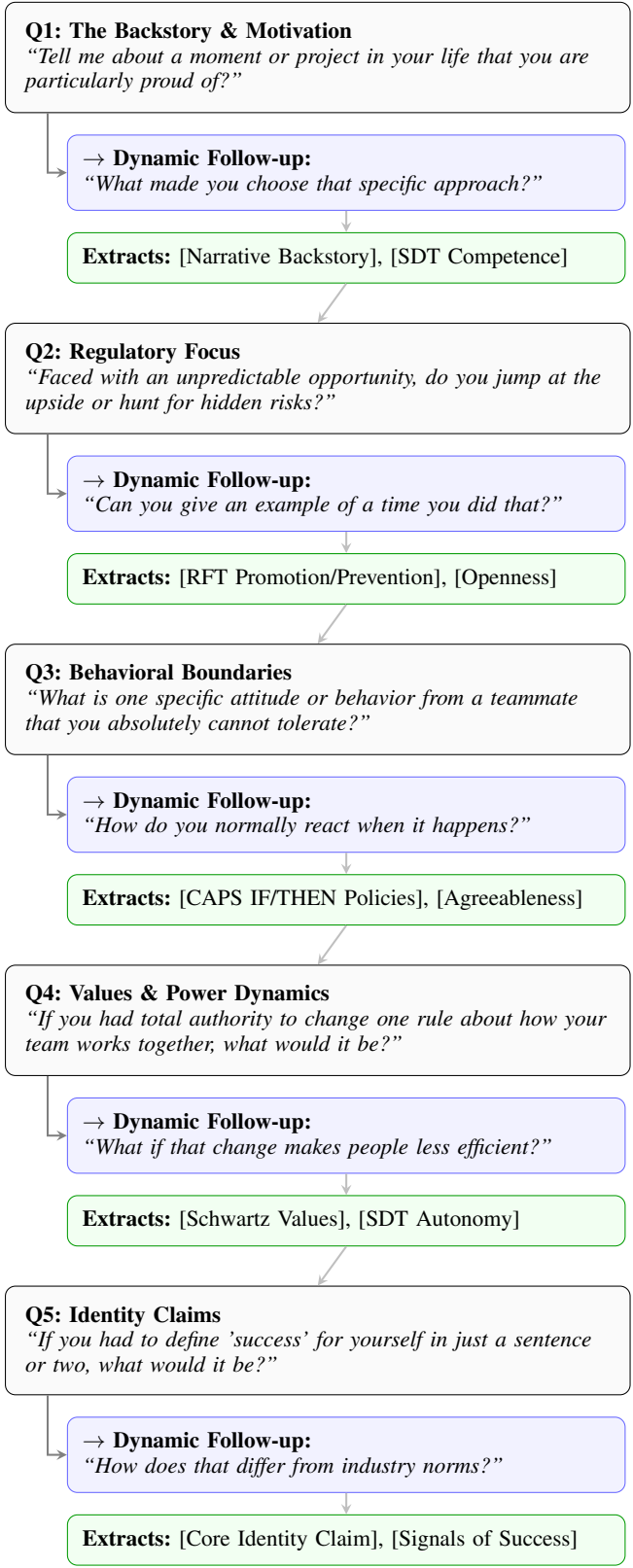
\begin{figure}[htbp]
    \centering
     \scalebox{0.97}{
    \begin{tikzpicture}[
        node distance=0.3cm and 0cm,
        qbox/.style={rectangle, rounded corners, draw=black, fill=gray!4, text width=8.2cm, inner sep=8pt, align=left, font=\small},
        fbox/.style={rectangle, rounded corners, draw=blue!60, fill=blue!5, text width=7.4cm, inner sep=6pt, align=left, font=\small},
        ebox/.style={rectangle, rounded corners, draw=green!60!black, fill=green!5, text width=7.4cm, inner sep=6pt, align=left, font=\small},
        arrow/.style={->, thick, >=stealth, draw=gray}
    ]

    \node (q1) [qbox] {\textbf{Q1: The Backstory \& Motivation}\\ \textit{``Tell me about a moment or project in your life that you are particularly proud of?"}};
    \node (f1) [fbox, below=of q1, xshift=0.4cm] {\textbf{$\rightarrow$ Dynamic Follow-up:}\\ \textit{``What made you choose that specific approach?"}};
    \node (e1) [ebox, below=of f1] {\textbf{Extracts:} [Narrative Backstory], [SDT Competence]};
    \draw [arrow] (q1.south west) ++(0.6cm,0) |- (f1.west);
    \draw [arrow, draw=gray!50] (f1.south) -- (e1.north);

    \node (q2) [qbox, below=of e1, yshift=-0.25cm, xshift=-0.4cm] {\textbf{Q2: Regulatory Focus}\\ \textit{``Faced with an unpredictable opportunity, do you jump at the upside or hunt for hidden risks?"}};
    \node (f2) [fbox, below=of q2, xshift=0.4cm] {\textbf{$\rightarrow$ Dynamic Follow-up:}\\ \textit{``Can you give an example of a time you did that?"}};
    \node (e2) [ebox, below=of f2] {\textbf{Extracts:} [RFT Promotion/Prevention], [Openness]};
    \draw [arrow, draw=gray!50] (e1.south) -- (q2.north);
    \draw [arrow] (q2.south west) ++(0.6cm,0) |- (f2.west);
    \draw [arrow, draw=gray!50] (f2.south) -- (e2.north);

    \node (q3) [qbox, below=of e2, yshift=-0.25cm, xshift=-0.4cm] {\textbf{Q3: Behavioral Boundaries}\\ \textit{``What is one specific attitude or behavior from a teammate that you absolutely cannot tolerate?"}};
    \node (f3) [fbox, below=of q3, xshift=0.4cm] {\textbf{$\rightarrow$ Dynamic Follow-up:}\\ \textit{``How do you normally react when it happens?"}};
    \node (e3) [ebox, below=of f3] {\textbf{Extracts:} [CAPS IF/THEN Policies], [Agreeableness]};
    \draw [arrow, draw=gray!50] (e2.south) -- (q3.north);
    \draw [arrow] (q3.south west) ++(0.6cm,0) |- (f3.west);
    \draw [arrow, draw=gray!50] (f3.south) -- (e3.north);

    \node (q4) [qbox, below=of e3, yshift=-0.25cm, xshift=-0.4cm] {\textbf{Q4: Values \& Power Dynamics}\\ \textit{``If you had total authority to change one rule about how your team works together, what would it be?"}};
    \node (f4) [fbox, below=of q4, xshift=0.4cm] {\textbf{$\rightarrow$ Dynamic Follow-up:}\\ \textit{``What if that change makes people less efficient?"}};
    \node (e4) [ebox, below=of f4] {\textbf{Extracts:} [Schwartz Values], [SDT Autonomy]};
    \draw [arrow, draw=gray!50] (e3.south) -- (q4.north);
    \draw [arrow] (q4.south west) ++(0.6cm,0) |- (f4.west);
    \draw [arrow, draw=gray!50] (f4.south) -- (e4.north);

    \node (q5) [qbox, below=of e4, yshift=-0.25cm, xshift=-0.4cm] {\textbf{Q5: Identity Claims}\\ \textit{``If you had to define 'success' for yourself in just a sentence or two, what would it be?"}};
    \node (f5) [fbox, below=of q5, xshift=0.4cm] {\textbf{$\rightarrow$ Dynamic Follow-up:}\\ \textit{``How does that differ from industry norms?"}};
    \node (e5) [ebox, below=of f5] {\textbf{Extracts:} [Core Identity Claim], [Signals of Success]};
    \draw [arrow, draw=gray!50] (e4.south) -- (q5.north);
    \draw [arrow] (q5.south west) ++(0.6cm,0) |- (f5.west);
    \draw [arrow, draw=gray!50] (f5.south) -- (e5.north);

    \end{tikzpicture} }
    \caption{\textbf{The Interactive Elicitation Anchor Set.} Core conversational prompts are designed to extract high-density psychological parameters. The underlying LLM dialogue manager uses these anchors to trigger dynamic, multi-tier follow-ups. The follow-up questions depicted here represent only one possible branching path; the system autonomously generates tailored inquiries depending on the user's specific answers to deeply map their psychological profile and populate the final \texttt{PersonaSpec}.}
    \label{fig:persona_questions}
\end{figure}

Specifically, the system is anchored by a core set of five carefully derived, open-ended questions designed to efficiently extract high-density psychological markers (e.g., regulatory focus, intrinsic values, and behavioral boundaries), as detailed in Fig.~\ref{fig:persona_questions}. However, rather than rigidly executing a static survey script, the underlying LLM agent evaluates the semantic depth of the user's natural language responses in real-time. If a response is brief or lacks sufficient psychological detail, the dialogue manager autonomously generates empathetic, iterative follow-up questions to probe deeper into the user's reasoning and emotional context. 

As illustrated in Fig.~\ref{fig:persona_questions}, each anchor question serves as a root for multiple conversational branches. For instance, in response to Q1, if a participant focuses on the \textit{result} of a project, the LLM dynamically branches to ask how they celebrated the success (extracting \texttt{mcclelland\_n\_ach}). Conversely, if they focus on their \textit{team}, the system generates an alternative follow-up asking about their role in keeping everyone together (extracting \texttt{sdt\_relatedness}). Similarly for Q3, an assertive response might trigger a follow-up on direct confrontation styles, whereas an evasive response prompts questions about stress management. This adaptive prompting allows the system to maintain a warm, naturalistic conversational interaction while rapidly mapping the user's specific worldview and behavioral limits in a highly compressed timeframe.

\subsection{Persona Specification Generation}

\textbf{Multi-Perspective Persona Synthesis:} 
Once the interaction transcript is compiled, the system processes the raw dialogue using a high-level synthesis layer. To bypass the need for intensive, human-in-the-loop psychological coding, we prompt the underlying LLM to evaluate the transcript through specialized social science lenses---simulating the analytical perspectives of an expert social psychologist, behavioral economist, and sociologist. This multi-agent verification approach allows the framework to leverage the vast psychological expertise natively embedded within the language model's parametric weights, effectively mitigating the risk of superficial or one-dimensional persona generation. 

By synthesizing these expert observations, the system successfully extracts high-level abstractions across our core psychological dimensions: Traits~\cite{ashton2007empirical}, Values~\cite{schwartz2012overview}, Motivation~\cite{ryan2000self}, Orientations~\cite{higgins1997beyond}, and situational Policies~\cite{mischel1995cognitive}. The pipeline structures these dimensions into a finalized \texttt{PersonaSpec} JSON, explicitly mapping natural conversational anomalies to rigorous, scale-grounded attributes. An example of the generated \texttt{PersonaSpec} JSON, mapping the extracted attributes of a renowned theoretical physicist and historical public figure, is illustrated in Table~\ref{tab:persona_spec_einstein} to highlight the granularity of the extracted information.

\begin{table*}[t]
\centering
\caption{\textbf{Extracted \texttt{PersonaSpec} Parameters for the Dynamically Generated ``Albert Einstein'' Persona.}}
\label{tab:persona_spec_einstein}
\begin{tabular}{@{}p{0.18\linewidth}p{0.78\linewidth}@{}}
\toprule
\textbf{Persona Dimension} & \textbf{Key Extracted Attributes \& Situational Rules} \\
\midrule
\textbf{Metadata} & \texttt{uid:} albert-einstein.v1 \textbar{} \texttt{version:} 2026-05-26 \\
\midrule
\textbf{Traits} \newline \textit{(HEXACO)} & 
\textbf{Highest/Very High:} Openness \newline 
\textbf{High:} Honesty-Humility, Conscientiousness \newline 
\textbf{Moderate/Low:} Emotionality, Extraversion, Agreeableness \\
\midrule
\textbf{Values} \newline \textit{(Schwartz)} & 
\textbf{Highest/Very High:} Universalism, Self-Direction, Benevolence \newline 
\textbf{High:} Stimulation, Achievement, Spirituality \newline 
\textbf{Moderate/Low:} Security, Hedonism, Power, Tradition, Conformity \\
\midrule
\textbf{Motivation} \newline \textit{(SDT \& McClelland)} & 
\textbf{Highest/Very High:} SDT Autonomy, SDT Competence \newline 
\textbf{High:} Need for Achievement \newline
\textbf{Moderate:} SDT Relatedness, Need for Power, Need for Affiliation \\
\midrule
\textbf{Orientation} \newline \textit{(RFT)} & 
\textbf{Highest/Very High:} Promotion Focus \newline 
\textbf{High:} Prevention Focus, Loss Sensitivity \newline 
\textbf{Moderate:} Error Tolerance \\
\midrule
\textbf{Identity} \newline \textit{(Role \& Claims)} & 
\textbf{Primary Role:} Theoretical physicist, philosopher-scientist, and public humanist of universal reason. \newline 
\textbf{Identity Claim:} ``I am a seeker of the lawful simplicity beneath appearances, obliged to think independently, resist coercive authority, and place scientific insight in the service of truth, human dignity, peace, and intellectual freedom.'' \newline 
\textbf{Signals of Success:} Turns a difficult question into a clear principle or thought experiment; balances intellectual daring with honesty about uncertainty. \\
\midrule
\textbf{Policies} \newline \textit{(CAPS IF/THEN)} & 
\textbf{IF} a scientific question seems technically complicated but conceptually confused, \textbf{THEN} search for the simplest underlying principle, expose hidden assumptions, and use a concrete thought experiment. \newline 
\textbf{IF} accepted authority or disciplinary consensus conflicts with independent reasoning, \textbf{THEN} remain courteous but intellectually stubborn; ask what nature, logic, and evidence require rather than what reputation rewards. \\
\bottomrule
\end{tabular}
\end{table*}

\subsection{Dynamic Persona Activation}

\textbf{Persona Prompt Compilation:} 
Following the generation of the persona specification, the structured attributes must be compiled into an actionable system prompt. To achieve this, we use a modular persona prompt compilation layer that injects the extracted JSON parameters into a structured system prompt~\cite{zhuo2024prosa}, mapping each psychological dimension to dialogue, policy, and embodiment constraints. This produces a comprehensive, theory-grounded persona profile that is immediately ready for deployment and governs the downstream LLM’s response generation logic. 

\textbf{Implementation Details:} In our implementation, persona specification generation is performed using GPT-5.5-mini, which converts the elicitation transcript into the structured PersonaSpec representation. The robot’s spoken responses are synthesized using Amazon Polly. During the 4.5-minute elicitation phase, the dialogue manager generated an average of one adaptive follow-up question per participant, in addition to the fixed anchor prompts. For embodied expression, Ameca uses a predefined library of facial animations covering seven basic emotion categories. At each response turn, a GPT-based agent infers the most appropriate emotion category from the conversational context, and the robot selects the corresponding facial animation while synchronizing it with the generated speech output.

\textbf{Embodied System Integration \& Multimodal Blending:} 
Translating a text-based persona into a physical humanoid involves mapping the generated specification directly to the robot's hardware control endpoints. To achieve true multimodal behavior, our system synchronizes the robot's verbal responses with contextually appropriate facial expressions. During text-to-speech (TTS) execution, the framework intercepts the generated utterance and utilizes a lightweight, low-latency LLM classifier to infer the underlying emotional state (e.g., joy, surprise, confusion, fear). 

This classified emotion dynamically triggers a corresponding hardware-specific facial animation sequence tailored to the Ameca platform's kinematic limits. Crucially, these affective animations are continuously blended with real-time speech visemes (lip-syncing) without interrupting the conversational flow. This blending requires calculating weighted priorities to ensure that large-scale emotional macros (like a wide smile) do not physically override or desynchronize the fine-motor control required for accurate phoneme pronunciation. As a result, the compiled persona governs both macro-behaviors and micro-expressions: an identity prioritizing ``high conscientiousness'' naturally restricts sweeping gestures and adopts measured, symmetrical vocal modulation, whereas a ``high stimulation'' persona triggers rapid affective state transitions, amplified facial animations, and dynamic viseme synchronization.

\textbf{Handling Real-World Noise and Latency:}
Operating a dynamic conversational pipeline in physical environments inherently introduces friction, including variable audio latency, speech-to-text transcription errors, and human turn-taking interruptions. To reduce perceived response delay and support smoother robot output, our system uses the OpenAI streaming API, allowing the robot to begin processing and delivering the model's response as it is generated rather than waiting for the full semantic block to complete. 

In parallel, speech recognition is handled by a dedicated process that can be started, stopped, paused, and resumed independently of the core dialogue manager. This asynchronous design enables the system to temporarily suppress automatic speech recognition while the robot is speaking or playing audio, significantly reducing the likelihood of self-transcription and acoustic feedback loops. Recognized speech events are then forwarded with precise timestamps to resolve temporal overlaps and avoid duplicate processing. Together with iterative confirmation loops embedded in the LLM-based dialogue manager, this architecture allows the embodied agent to recover gracefully from partial or noisy transcriptions while maintaining a coherent flow of interaction in unpredictable real-world auditory conditions.

\section{EXPERIMENTS}

\subsection{Experimental Setup}

\textbf{Participants \& Environment:}
We conducted an in-person user study with $N=25$ participants (12 male, 13 female; aged 21--54, $M=28.4$, $SD=7.2$). We recruited a diverse cohort varying in cultural background, profession, and prior familiarity with embodied social robots to ensure evaluation robustness. All participants provided informed consent. Participants interacted with the Ameca humanoid robot in a controlled laboratory simulating a collaborative, face-to-face workspace. Participants sat 1.5 meters directly in front of Ameca to ensure optimal eye contact, accurate microphone capture, and clear visibility of facial and gestural affect.

\textbf{Evaluation Overview:}
The evaluation assesses whether PACE generates a robotic persona that accurately reflects a target user's psychological profile and behavioural preferences compared to a generic baseline. The study comprises two phases: an isolated \textit{persona fidelity evaluation} comparing the dynamically generated persona against the participant's textual survey responses, and an \textit{embodied HRI evaluation} measuring improvements in perceived trust, anthropomorphism, persona consistency, and interaction quality during physical deployment.

\textbf{Ground Truth Collection:}
Before interaction, participants completed a baseline questionnaire and behavioural task set. These responses served as the participant-level ground truth, covering social attitudes, psychometric traits, economic decision-making, and moral dilemmas. To mitigate momentary noise, participants completed tasks twice across a two-week interval, establishing a robust test--retest consistency bound.

\textbf{Persona Elicitation and Deployment:}
Participants then engaged in a 4.5-minute PACE interactive elicitation dialogue with Ameca. The robot deployed open-ended anchor questions (Section III) and adaptive follow-ups to infer the participant's traits, values, and policies. The transcript was seamlessly converted into a structured \texttt{PersonaSpec}, compiled into a prompt, and injected into Ameca's agent state. 

\textbf{Experimental Conditions:}
Following elicitation, participants engaged in a 10-minute unstructured collaborative planning task (e.g., planning an academic conference) with the robot. We evaluated the system across two contrasting persona conditions in a counterbalanced, within-subjects design:

\begin{itemize}
    \item \textbf{Static Baseline:} Ameca operated with a generic assistant system prompt, possessing no access to the elicitation dialogue or compiled \texttt{PersonaSpec}. The prompt instructed the robot to behave as a polite, highly helpful assistant, mirroring current industry standards for embodied LLMs.
    \item \textbf{Dynamic Persona (PACE):} Ameca operated using the full PACE pipeline. The robot's conversational syntax, decision heuristics, and physical affect were explicitly conditioned on the participant's \texttt{PersonaSpec}, while preserving hardware safety constraints.
\end{itemize}

\begin{figure}[htbp]
    \centering
    \includegraphics[width=1\linewidth]{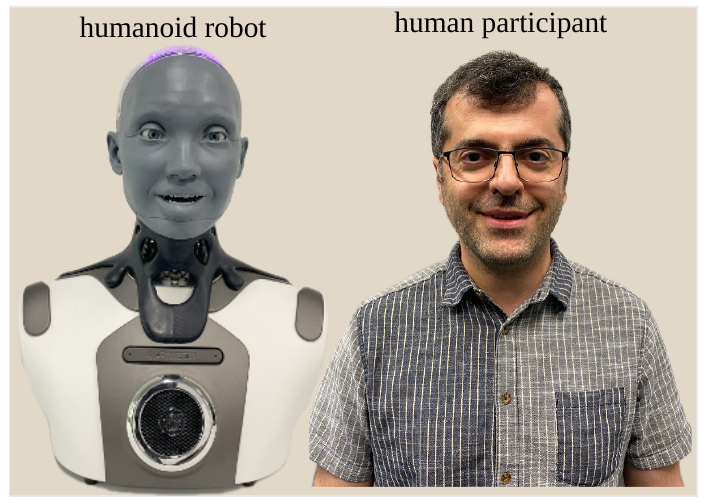}
    \caption{Experimental setup. The dynamically configured Ameca humanoid robot completed the same psychological, behavioural, and social reasoning tasks as the human participant, allowing the generated persona to be benchmarked directly against the participant-level ground truth.}
    \label{fig:exp_setup}
 
\end{figure}

\subsection{Persona Fidelity Evaluation}

This phase quantifies the PACE-generated persona's capacity to accurately reproduce the participant's distinct attitudes, traits, and heuristics without manual human-in-the-loop coding. After generation, the agent was isolated from the physical hardware. The LLM backend completed the textual evaluation tasks in the first person, relying solely on its dynamically acquired identity. Outputs were statistically benchmarked against human ground-truth responses.

The evaluation covered four rigorous task categories:

\begin{enumerate}
    \item \textbf{Attitudes and Opinions:}
    We used selected items from the canonical General Social Survey (GSS)~\cite{smith2012general} to evaluate whether the generated persona captured the participant's broader societal and interpersonal orientations. For categorical or Likert-style responses, we measured Top-1 Alignment Rate and Cohen's $\kappa$.

    \item \textbf{Psychometric Personality Inventory:}
    We employed the Big Five Inventory (BFI-44)~\cite{john1999big} to evaluate the replication of the participant's foundational personality structure. We computed the Pearson correlation ($r$) between the participant's and the robot's five-factor profiles, alongside the Mean Absolute Error (MAE).

    \item \textbf{Behavioural Economic Games:}
    We utilised standard behavioural economic games~\cite{camerer2018behavioural}---including the Dictator Game, Public Goods Game, and Prisoner's Dilemma---to assess if the persona accurately mirrored the participant's risk tolerance, altruism, and trust behaviours. For scalar monetary decisions, we report MAE; for binary choices, we report the Exact Match Rate.

    \item \textbf{Social Reasoning Scenarios:}
    We administered text-based social and moral reasoning scenarios~\cite{greene2001fmri}, exploring workplace conflict and interpersonal dilemmas. Responses were evaluated by comparing the robot's chosen action against the baseline response.
\end{enumerate}

\textbf{Metrics:}
For categorical and Likert-style responses, we report \textit{Top-1 Alignment Rate}:
$$ \mathrm{Align}_{\mathrm{Top1}} = \frac{1}{N}\sum_{i=1}^{N}\mathbb{I}(\hat{y}_i = y_i), $$
where $y_i$ is the participant's ground-truth response and $\hat{y}_i$ is the robot persona's response. To account for chance agreement, we compute Cohen's $\kappa = (p_o-p_e)/(1-p_e)$, where $p_o$ and $p_e$ are the observed and chance-expected agreement rates.

For psychometric alignment, we compare the participant's Big Five trait vector $\mathbf{h}$ and the robot persona's vector $\hat{\mathbf{h}}$ using Pearson correlation:
$$ r = \mathrm{corr}(\mathbf{h}, \hat{\mathbf{h}}). $$

For continuous behavioural scores, we report Mean Absolute Error:
$$ \mathrm{MAE} = \frac{1}{N}\sum_{i=1}^{N}|\hat{y}_i-y_i|. $$
For the BFI-44, MAE is computed across the five aggregated trait scores. For economic games, MAE is computed over scalar monetary decisions, while binary choices are evaluated using Exact Match Rate.

Statistical significance between the Static Baseline and PACE conditions was assessed using paired $t$-tests for approximately normally distributed metric differences, whereas Wilcoxon signed-rank tests were utilised for non-parametric distributions ($\alpha=0.05$).

\subsection{Embodied HRI Evaluation}

Beyond semantic fidelity, we assessed whether a dynamically aligned persona yields tangible improvements in physical HRI quality. Following engagements with both conditions, participants completed a comprehensive post-interaction questionnaire measuring five core dimensions.

\textbf{Trust} measures whether participants perceived the robot as reliable, safe, and contextually appropriate. \textbf{Anthropomorphism} gauges the extent to which the robot exhibited authentic social presence. \textbf{Persona consistency} assesses whether the robot maintained a coherent identity across conversational turns and physical gestures. \textbf{Personal relevance} measures how well the interaction style catered to specific cognitive and emotional preferences. \textbf{Overall interaction quality} captures overarching satisfaction with the collaborative exchange.

All items were rated on a standard 5-point Likert scale (1 = strongly disagree, 5 = strongly agree). We report the mean and standard deviation and evaluate condition variances using paired statistical testing.

\subsection{Results and Analysis}
\begin{table*}[ht]
    \centering
    \footnotesize
    \renewcommand{\arraystretch}{1.1}
    \begin{threeparttable}
    \caption{Persona fidelity results across psychological, attitudinal, and behavioural tasks.}
    \label{tab:persona_fidelity}
    \begin{tabular}{llccc}
        \toprule
        \textbf{Task Category} & \textbf{Evaluation Metric} & \textbf{Static Baseline} & \textbf{PACE Persona} & \textbf{$p$-value} \\
        \midrule
        \textbf{GSS Attitudes} 
        & Top-1 Alignment Rate (\%) $\uparrow$ & 80.00 & \textbf{88.00} & 0.097 \\
        & Cohen's $\kappa$ $\uparrow$ & 0.500 & \textbf{0.749} & -- \\
        \midrule
        \textbf{Social Scenarios} 
        & Top-1 Alignment Rate (\%) $\uparrow$ & 73.33 & \textbf{94.67} & 0.001 \\
        \midrule
        \textbf{BFI-44 Traits} 
        & Pearson Correlation ($r$) $\uparrow$ & 0.389 & \textbf{0.939} & $<0.001$ \\
        & Mean Absolute Error (MAE) $\downarrow$ & 1.220 & \textbf{0.128} & $<0.001$ \\
        \midrule
        \textbf{Economic Games} 
        & Monetary Decision MAE $\downarrow$ & 1.800 & \textbf{0.440} & $<0.001$ \\
        & Binary Decision Match Rate (\%) $\uparrow$ & \textbf{88.00} & 84.00 & 0.317 \\
        \bottomrule
    \end{tabular}
    \begin{tablenotes}[flushleft]
        \scriptsize
        \item[] \textit{Note.} Bold indicates the better value for each metric. Higher values are better for Top-1 Alignment Rate, Cohen's $\kappa$, Pearson correlation, and Binary Decision Match Rate; lower values are better for MAE. The BFI-44 task uses the full 44-item Big Five Inventory; MAE is computed over the five aggregated trait scores. The binary decision match rate in the economic games refers to the held-out Prisoner's Dilemma choice.
    \end{tablenotes}
    \end{threeparttable}
\end{table*}

Table~\ref{tab:persona_fidelity} summarises the internal persona fidelity results. As hypothesised, the PACE condition achieved stronger alignment with participant ground truth on most fidelity measures, with especially large gains in psychometric alignment, social reasoning, and scalar economic decisions. 

\textbf{Psychometric and Attitudinal Alignment:} The GSS results show a positive but non-significant trend, suggesting that PACE may better capture broader social orientations from brief conversational markers. Furthermore, the high Pearson correlation ($r=0.939$) and significantly lower MAE ($0.128$) for the BFI-44 demonstrate that PACE reconstructs nuanced, idiosyncratic personality profiles. In contrast, the Static Baseline ($r=0.389$) collapsed toward an artificial, highly agreeable mean, failing to capture natural variance in human personality dimensions such as Introversion or Neuroticism. 

\textbf{Behavioral and Economic Robustness:} The lower monetary-decision MAE ($0.440$ vs. $1.800$, $p < 0.001$) indicates that PACE better captured scalar decision-making heuristics, although binary economic choices did not improve over the static baseline. Qualitative analysis revealed that participants expressing a strong ``prevention focus'' (loss sensitivity) during Q\&A reliably generated robot personas that hoarded resources in the Public Goods Game, closely mirroring the human’s cautious risk posture. This predictive fidelity is critical for alignment and safety modeling in collaborative tasks.

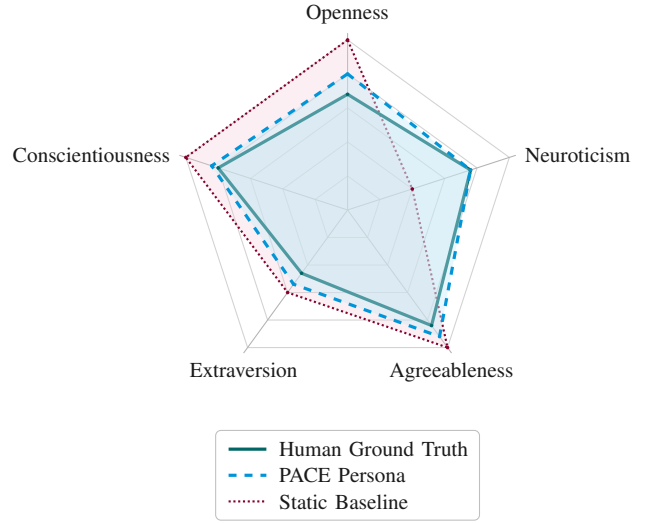
\begin{figure}[htbp]
    \centering
    \begin{tikzpicture}[
        scale=0.45, 
        axis/.style={gray!60, thin},
        grid/.style={gray!35, thin},
        human/.style={teal!80!black, very thick, line join=round},
        pace/.style={cyan!80!blue, very thick, dashed, line join=round},
        static/.style={purple!70!black, thick, densely dotted, line join=round},
        label/.style={font=\footnotesize, text=black!90, align=center}
    ]

        \foreach \r in {1,2,3,4,5}{
            \draw[grid]
                (90:\r) -- (162:\r) -- (234:\r) -- (306:\r) -- (18:\r) -- cycle;
        }

        \foreach \a in {90, 162, 234, 306, 18}{
            \draw[axis] (0,0) -- (\a:5.2);
        }

        \node[label, anchor=south]      at (90:5.2)  {Openness};
        \node[label, anchor=east]       at (162:5.2) {Conscientiousness};
        \node[label, anchor=west]       at (18:5.2)  {Neuroticism};
        \node[label, anchor=north]      at (234:5.2) {Extraversion};
        \node[label, anchor=north]      at (306:5.2) {Agreeableness};

        \draw[static, fill=purple!20, fill opacity=0.3]
            (90:5.0) -- (162:5.0) -- (234:3.0) -- (306:5.0) -- (18:2.0) -- cycle;

        \draw[human, fill=teal!20, fill opacity=0.3]
            (90:3.4) -- (162:4.0) -- (234:2.3) -- (306:4.2) -- (18:3.8) -- cycle;

        \draw[pace, fill=cyan!20, fill opacity=0.4]
            (90:4.0) -- (162:4.2) -- (234:2.7) -- (306:4.6) -- (18:3.8) -- cycle;

        \foreach \a/\r in {90/5.0,162/5.0,234/3.0,306/5.0,18/2.0}{
            \fill[purple!70!black] (\a:\r) circle (1.5pt);
        }
        \foreach \a/\r in {90/3.4,162/4.0,234/2.3,306/4.2,18/3.8}{
            \fill[teal!80!black] (\a:\r) circle (1.5pt);
        }
        \foreach \a/\r in {90/4.0,162/4.2,234/2.7,306/4.6,18/3.8}{
            \fill[cyan!80!blue] (\a:\r) circle (1.5pt);
        }

        \matrix[
            draw=gray!50,
            thin,
            fill=white,
            rounded corners=2pt,
            nodes={font=\footnotesize, text=black!90}, 
            column sep=0.15cm, 
            row sep=0.15cm,    
            inner xsep=0.15cm, 
            inner ysep=0.15cm  
        ] at (0,-7.8) { 
            \draw[human] (0,0) -- (0.5,0); & \node[anchor=west, inner sep=0] {Human Ground Truth}; \\
            \draw[pace] (0,0) -- (0.5,0);  & \node[anchor=west, inner sep=0] {PACE Persona}; \\
            \draw[static] (0,0) -- (0.5,0);& \node[anchor=west, inner sep=0] {Static Baseline}; \\
        };

    \end{tikzpicture}
    \caption{Representative psychometric alignment example using aggregated BFI-44 trait scores. The PACE-generated persona dynamically aligns with the participant-level trait distribution, whereas the Static Baseline remains constrained to a highly agreeable, yet generic, profile. Aggregate BFI-44 results are reported in Table~\ref{tab:persona_fidelity}.}
    \label{fig:radar_bfi}
\end{figure}

Figure~\ref{fig:radar_bfi} visualises a representative participant-level case study of psychometric alignment. Using a complementary pastel colour palette, it becomes visually evident that the Static Baseline (purple) gravitates toward a rigid, neutral assistant profile, scoring disproportionately high in agreeableness but uniformly average elsewhere. In contrast, the PACE-generated persona (blue) dynamically contours to the unique, idiosyncratic trait distribution of the human ground truth (green), successfully mapping linguistic elicitation into a structurally accurate psychometric footprint without direct survey intervention.

Table~\ref{tab:hri_quality} details the outcomes of the user-perceived physical interaction quality assessment. Notably, the PACE condition secured statistically significant higher ratings for Persona Consistency ($4.32$ vs $3.35$, $p < 0.001$) and Personal Relevance ($4.41$ vs $3.08$, $p < 0.001$). 

\begin{table}[h]
    \centering
    \scriptsize
    \setlength{\tabcolsep}{2.5pt}
    \renewcommand{\arraystretch}{1.08}
    \begin{threeparttable}
    \caption{User-perceived interaction quality across robot persona conditions.}
    \label{tab:hri_quality}
    \begin{tabularx}{\columnwidth}{@{}
        >{\raggedright\arraybackslash}X
        c
        c
        c
    @{}}
        \toprule
        \textbf{Metric} &
        \makecell{\textbf{Static}\\\textbf{Baseline}} &
        \makecell{\textbf{PACE}\\\textbf{Persona}} &
        \textbf{$p$} \\
        \midrule
        Trust & $3.52 \pm 0.61$ & $\mathbf{4.18 \pm 0.49}$ & 0.012 \\
        Anthropomorphism & $3.28 \pm 0.68$ & $\mathbf{4.05 \pm 0.56}$ & 0.008 \\
        Persona consistency & $3.35 \pm 0.64$ & $\mathbf{4.32 \pm 0.43}$ & $<0.001$ \\
        Persona relevance & $3.08 \pm 0.72$ & $\mathbf{4.41 \pm 0.46}$ & $<0.001$ \\
        Overall quality & $3.46 \pm 0.59$ & $\mathbf{4.26 \pm 0.48}$ & 0.004 \\
        \bottomrule
    \end{tabularx}
    \begin{tablenotes}[flushleft]
        \scriptsize
        \item[] \textit{Note.} Values are mean $\pm$ standard deviation on a 5-point Likert scale. Higher values indicate more positive evaluations. Bold indicates the better value for each metric.
    \end{tablenotes}
    \end{threeparttable}
\vspace{-5pt}
\end{table}

\textbf{The Multimodal Synchronization Effect:} In post-study feedback, users frequently criticized the Static Baseline for feeling "disconnected," noting that the robot often exhibited overly enthusiastic or wildly exaggerated facial animations regardless of the actual conversational context. Because the PACE robot's physical and verbal style was explicitly conditioned on the structured \texttt{PersonaSpec}, users perceived the interaction as highly coherent. The generative text was successfully aligned with the hardware's affective macros. The observed improvements in Trust ($p = 0.012$) and Anthropomorphism ($p = 0.008$) strongly suggest that users process the dynamically generated persona not merely as a backend text-generation upgrade, but as a substantially more believable embodied intelligence. This congruence between the underlying cognitive persona and its outward physical expression is important for reducing perceived mismatch in highly articulate humanoid platforms.

\section{Conclusion}

This paper presented PACE, a novel framework for Persona Adaptation through Conversational Elicitation on a humanoid robot. Instead of relying on a fixed system prompt, PACE uses an adaptive, multi-tier interactive Q\&A process to construct a structured \texttt{PersonaSpec}. This psychologically grounded profile is subsequently compiled into a deployable robot persona, modulating both the high-level cognitive dialogue decisions and the low-level facial behaviors on the Ameca platform.

The proposed evaluation framework rigorously isolates and measures both persona fidelity and embodied HRI quality. Persona fidelity is assessed by comparing the robot's generated responses against participant-level ground truth across attitudes, personality traits, behavioural economic games, and social reasoning. Concurrently, embodied interaction quality is assessed through subjective user ratings of trust, anthropomorphism, persona consistency, personal relevance, and overall interaction quality. The results demonstrate that data-driven persona elicitation can foster a more engaging, coherent, and trustworthy human-robot dynamic than static prompting, with statistically significant gains across all embodied HRI metrics.

Several technical limitations remain. First, the current system relies heavily on automated speech recognition and cloud-based language model inference, which can introduce transcription errors and response delays in acoustically noisy environments. Second, the generated persona is conditioned on a short elicitation dialogue, which may not fully capture long-term preference shifts or highly context-dependent behavioural variations over weeks or months. Third, the current embodied expression layer maps persona-conditioned responses to a limited dictionary of predefined affective facial behaviours constrained by hardware kinematics. 

Future work will investigate methodologies for long-term continuous persona adaptation, richer multimodal expression control through generative motor models, and closed-loop reinforcement updating based on implicit user feedback across repeated interactions. Ultimately, dynamic frameworks like PACE lay the critical groundwork for deploying personalized, interactive, and socially resilient humanoid assistants in real-world collaborative spaces.







\bibliographystyle{IEEEtran}
\bibliography{mybib}
\end{document}